\documentclass[conference]{IEEEtran}
\IEEEoverridecommandlockouts
\usepackage[
pdfstartview=FitH,
CJKbookmarks=true,
bookmarksnumbered=true,
bookmarksopen=true,
colorlinks, 
pdfborder=001,   
linkcolor=black,
anchorcolor=black,
citecolor=black,
urlcolor=black
]{hyperref} 
\usepackage{cite}
\usepackage{amsmath,amssymb,amsfonts}
\usepackage{algorithmic}
\usepackage{graphicx}
\usepackage{textcomp}
\usepackage{xcolor}

\def\BibTeX{{\rm B\kern-.05em{\sc i\kern-.025em b}\kern-.08em
    T\kern-.1667em\lower.7ex\hbox{E}\kern-.125emX}}
\begin{document}

\title{A Benchmark for Multi-UAV Task Assignment of an Extended Team Orienteering Problem}

\author{\IEEEauthorblockN{1\textsuperscript{st} Kun Xiao}
\IEEEauthorblockA{\textit{Beijing Institute of Aerospace Systems Engineering} \\
Beijing, China \\
robin\_shaun@foxmail.com}
\\
\\
\IEEEauthorblockN{3\textsuperscript{rd} Ying Nie}
\IEEEauthorblockA{\textit{Beijing Aerospace Automatic Control Institute} \\
Beijing, China \\
792761529@qq.com}
\\
\\
\IEEEauthorblockN{5\textsuperscript{th} Xiangke Wang}
\IEEEauthorblockA{\textit{College of Intelligence Science and Technology} \\
\textit{National University of Defense Technology}\\
Changsha, China \\
xkwang@nudt.edu.cn}
\and
\IEEEauthorblockN{2\textsuperscript{nd} Junqi Lu}
\IEEEauthorblockA{\textit{College of Intelligence Science and Technology} \\
\textit{National University of Defense Technology}\\
Changsha, China \\
970645730@qq.com}
\\
\IEEEauthorblockN{4\textsuperscript{th} Lan Ma}
\IEEEauthorblockA{\textit{College of Intelligence Science and Technology} \\
\textit{National University of Defense Technology}\\
Changsha, China\\
malan9608@163.com}
\\
\IEEEauthorblockN{6\textsuperscript{th} Guohui Wang}
\IEEEauthorblockA{\textit{China Academy of Launch Vehicle Technology} \\
Beijing, China \\
708869729@qq.com}
}

\maketitle

\begin{abstract}
A benchmark for multi-UAV task assignment is presented in order to evaluate different algorithms. An extended Team Orienteering Problem is modeled for a kind of multi-UAV task assignment problem. Three intelligent algorithms, i.e., Genetic Algorithm, Ant Colony Optimization and Particle Swarm Optimization are implemented to solve the problem. A series of experiments with different settings are conducted to evaluate three algorithms. The modeled problem and the evaluation results constitute a benchmark, which can be used to evaluate other algorithms used for multi-UAV task assignment problems.
\end{abstract}


\section{Introduction}
Unmanned aerial vehicles (UAVs) develop rapidly due to their large potential in both civilian and military uses, such as disaster rescue, reconnaissance and surveillance. Limited by its size and capability, a single UAV can hardly complete complex and persistent tasks\cite{b1}. Therefore, swarms of UAVs are emerging as a disruptive technology to enable highly-reconfigurable, on-demand, distributed intelligent autonomous systems with high impact on many areas of science, technology, and society\cite{b2}.

To achieve cooperation between UAVs, task assignment is necessary to make them conduct tasks in a good order and maximize total performance. The basic task assignment problem can be formulated as a Vehicle Routing Problem (VRP)\cite{b3}. VRP asks what the optimal set of routes for a fleet of vehicles is to traverse in order to deliver to a given set of customers. For VRP, all the targets need to be reached and no time limit is set, which is unsuitable for many kinds of task assignment problems. Compared with VRP, Team Orienteering Problem (TOP) considers time limit and its goal is to maximize total reward under the time limit \cite{b4}. Conventional TOP considers all vehicles have the same speed, which is unsuitable for the heterogeneous UAV swarm. And it doesn't consider the time cost when the UAV executes the task after reaching the target. To solve the unsuitability, we extend TOP, in which different UAVs have different flight speeds and different targets have different time costs. Moreover, unlike VRP and TOP, each UAV is unnecessary to come back to the depot in our proposed problem. The object of our proposed problem is to obtain as more reward as possible under certain time limit.   

The extended TOP is suitable for a wide range of multi-UAV task assignment problems, such as reconnaissance and transportation. Therefore, it can be a benchmark to evaluate different algorithm. In this paper, three intelligent algorithms, Genetic Algorithm (GA) , Ant Colony Optimization (ACO) and Particle Swarm Optimization (PSO) are tested under a series of experiments. The experiment environment, settings and analysis, together with the implementation of three algorithms are open sourced \footnote{Source code at \url{https://gitee.com/robin_shaun/multi-uav-task-assignment-benchmark}\\or \url {https://github.com/robin-shaun/Multi-UAV-Task-Assignment-Benchmark}}. Researchers can use the benchmark to evaluate their own algorithms.

\section{Problem Formation}

The extended TOP is built on a directed graph. A complete graph $G=(V,A)$ is given, where $V=\{0,...,n\}$ is the set of vertices and $A$ is the set of arcs. Vertices in $N=V \backslash\{0\}=\{1,...,n\}$ correspond to the targets, and vertex $0$ corresponds to the depot where UAVs start. $d_{ij}$ is the distance from vertex $i\in V$ to vertex $j\in V$ and $d_{ij}=d_{ji}$. $r_i$ is the reward associated with target $i$ and $r_i>0$ when $i \neq 0$ while $r_0=0$ because the depot cannot supply any reward. $t_i$ is the time consumption to finish the mission at target $i$. $T_{max}$ is the time limit of the total task. If a UAV arrives target $i$ but the remaining time is less than $t_i$, it cannot obtain the reward $r_i$.

Given a set of $K$ of UAVs, the TOP calls for the determination of at most $|K|$ UAV routes that maximize the total collected reward, while satisfying a maximum duration constraint \cite{b5}. The extended TOP has the same goal with TOP. $y_{i,k}$ is binary variable equal to 1 if target $i\in V$ is visited by UAV $k \in K$, and $0$ otherwise. $x_{ijk}$ is binary variable equal to 1 if path $(i, j) \in A$ is traversed by UAV $k$, and $0$ otherwise. $s_k$ is the flight speed of UAV $k$.

The mathematical programming formulation for the extended TOP is as follows.

$$
\text { maximize } \sum_{i \in V} r_{i} \sum_{k \in K} y_{i k}\\
$$
$$
\begin{array}{ll}
\text{ s.t. } &\sum_{j \in V} x_{i j k}=y_{i k} \quad \forall i \in V, k \in K \\
&\sum_{j \in V} x_{j i k}=y_{i k} \quad \forall i \in V, k \in K \\
&\sum_{k \in K} y_{0 k} \leq|K| \\
&\sum_{k \in K} y_{i k} \leq 1 \quad i \in V \backslash\{0\} \\
&\sum_{(i, j) \in \delta^{+}(S)} x_{i j k} \geq y_{b k} \quad \forall S \subseteq V \backslash\{0\}, b \in S, k \in K \\
&\sum_{(i, j) \in A} \frac{d_{i j}}{s_k} x_{i j k}+t_i y_{i k} \leq T_{\max } \quad \forall k \in K \\
&y_{i k} \in\{0,1\} \quad \forall i \in V, k \in K \\
&x_{i j k} \in\{0,1\} \quad \forall(i, j) \in A, k \in K
\end{array}
$$

\begin{figure}[htbp]
	\centerline{\includegraphics[width=9cm]{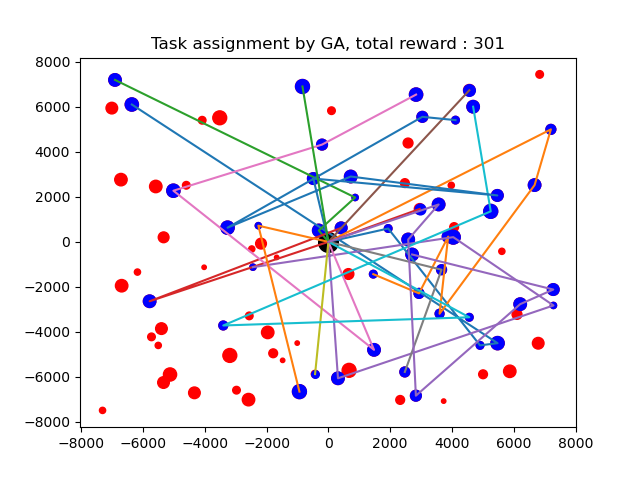}}
	\caption{The extended TOP solved by GA}
	\label{fig1}
\end{figure}

Even though the position coordinate system is unnecessary for the problem, it is built to visualize the result. Fig.~\ref{fig1} shows the extended TOP solved by GA. The red points are the targets not reached and the blue points are the targets reached. The black vertex is the depot. The size of the point is proportional to the reward. Lines with different colors are paths traversed by different UAVs. 

\section{Design of Three Intelligent Algorithms}

In this section, three intelligent algorithms, Genetic Algorithm, Ant Colony Optimization and Particle Swarm Optimization are designed to solve the extended TOP.

\subsection{Genetic Algorithm}

Genetic algorithm (GA) is a method to search the optimal solution by simulating natural selection and genetic mechanism of biological evolution process\cite{b6}. The algorithm transforms the process of solving a searching problem into a process similar to the crossover and mutation of chromosome during biological evolution. While dealing with complex combination optimization problems with large solution space, genetic algorithm can obtain great results quickly.

\begin{figure}[htbp]
	\centerline{\includegraphics[width=9cm]{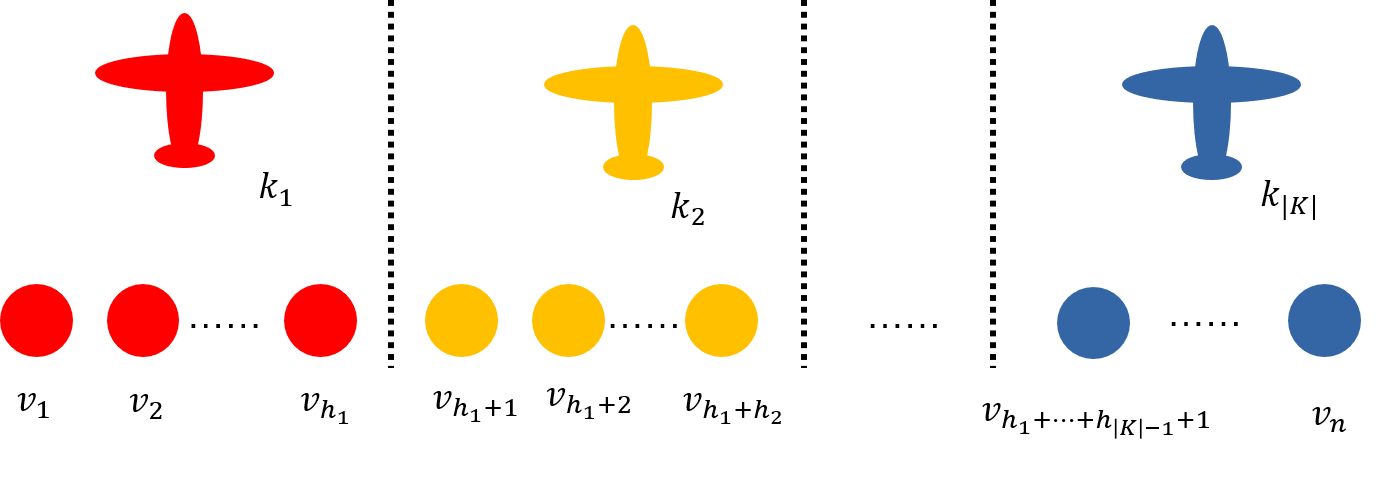}}
	\caption{Genetic representation of the solution domain}
	\label{fig2}
\end{figure}

The first step is to determine a genetic representation of the solution domain and a fitness function to evaluate the solution domain. Assuming that the time limit is large enough so that all target can be reached, we can determine a string $\epsilon$ by arranging all the targets\cite{b7}. The length of string $\epsilon$ is equal to the total number of targets. And then, we can determine a string $\delta$ by dividing string $\epsilon$ into $|K|$ groups\cite{b8}. The length of string $\delta$ is $|K|-1$. The combination of string $\epsilon$ and string $\delta$ corresponds to a feasible solution. Fig.~\ref{fig2} shows the genetic representation. The fitness function is defined as the total reward.

\begin{figure}[htbp]
	\centerline{\includegraphics[width=5cm]{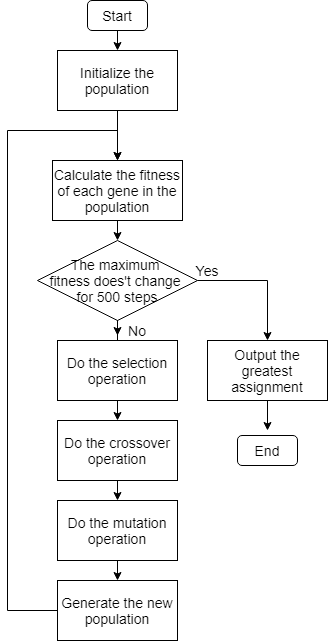}}
	\caption{Flow chart of GA}
	\label{fig3}
\end{figure}

The flow chart of GA is shown as Fig.~\ref{fig3}. In the selection operation, the roulette is performed on the new population combined by the parent population and offspring population to generate a new parent population. In the crossover operation, any two gene codes in the new parent population exchanges their codes with each other at a rate of 0.6. In the mutation operation, each code in the population changed in its value range at a rate of 0.05. After the crossover operation and the mutation operation, a new offspring population is generated. In order to speed up the convergence of genetic algorithm, the termination condition is set as whether the maximum fitness of the population does not change for 500 steps.

\subsection{Ant Colony Optimization}

\begin{figure}[htbp]
	\centerline{\includegraphics[width=9cm]{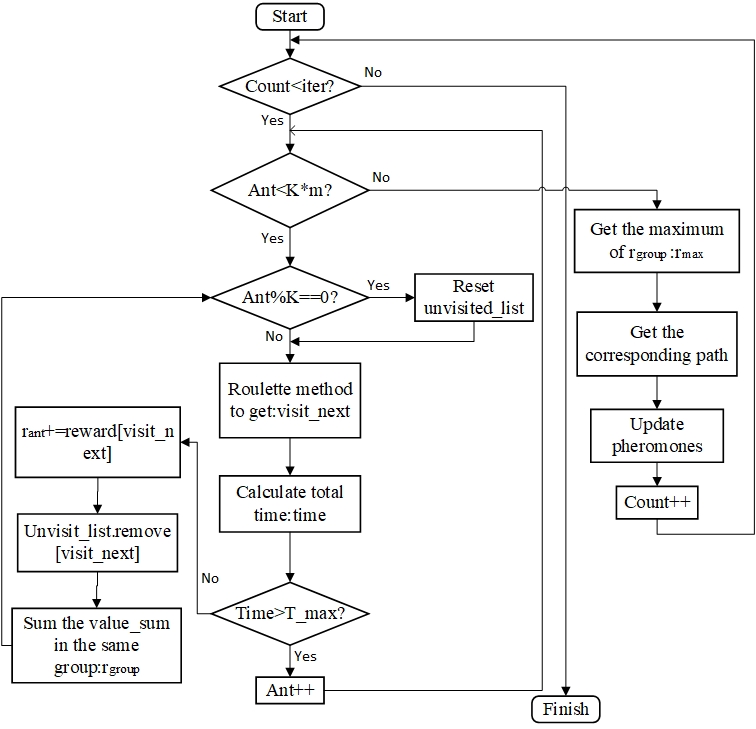}}
	\caption{Flow Chart of ACO}
	\label{fig4}
\end{figure}

The idea of ant colony optimization (ACO) is firstly given in 1989\cite{b9}, and gradually implemented as a probabilistic technique for solving computational problems which can be reduced to finding good paths through graphs\cite{b10}. Currently, the great majority of problems attacked by ACO are which all the necessary information is available and does not change during problem solution\cite{b11}. Hence, it is a great method for solving this problem. The flow chart of ACO is shown as Fig.~\ref{fig4}.

The ants in the ant colony are equally divided into $m$ groups. Since there are $|K|$ UAVs (with different speeds), the number of ants in each group is set as $|K|$. In other words, there are $|K|$ types of ants. The target points of each group of ants are not repeated, so the unvisited list would be reset only when a group of ants are traversed.

The next target of each ant could be obtained by roulette method. The reward function used for evaluating the solution is defined as the sum of reward obtained by all ants in the group, denoted by $r_{group}$. And the reward function used for evaluating each ant is defined as the sum of the reward obtained by the ant, denoted by $r_{ant}$. And $r_{max}$ is the maximum of all the $r_{group}$. Because of the time limit, the heuristic function should be not only positively related to value, but also negatively related to time. Thus, the heuristic function $H$ is designed as

$$
H(\text {ant}, j)=\frac{s_{\text {ont}} \times r_{j}}{d_{j-1 j} \times t_{j}}
$$

where $j \in V$

The number of iterations is set as a constant $iter$. The rewards of a group is related to the ants in the group, while pheromone of a type is related to the ants belonging to the type. The total number of ants in an iteration is $|K|\times m$, so tremendous number of ants are needed for solving the problem. In order to improve the convergence speed, the volatilization factor(V) of each type of pheromone is determined by the reward obtained by the type of ants in one iteration. 

$$
\mathrm{V}(\text{type})=\sum_{\text {ant } \in \text { type }} \frac{r_{\text {ant }}}{1+\left(r_{\text {max }}-r_{\text {group }}\right)^{\eta}} /m
$$

\subsection{Particle Swarm Optimization}

Particle swarm optimization (PSO) is a global random search algorithm which simulates the migration and swarm behavior of birds in the process of foraging. Its basic core is to make use of the information shared by the individuals in the group, so that the movement of the whole group will evolve from disorder to order in the problem solving space \cite{b12}.

\begin{figure}[htbp]
	\centerline{\includegraphics[width=9cm]{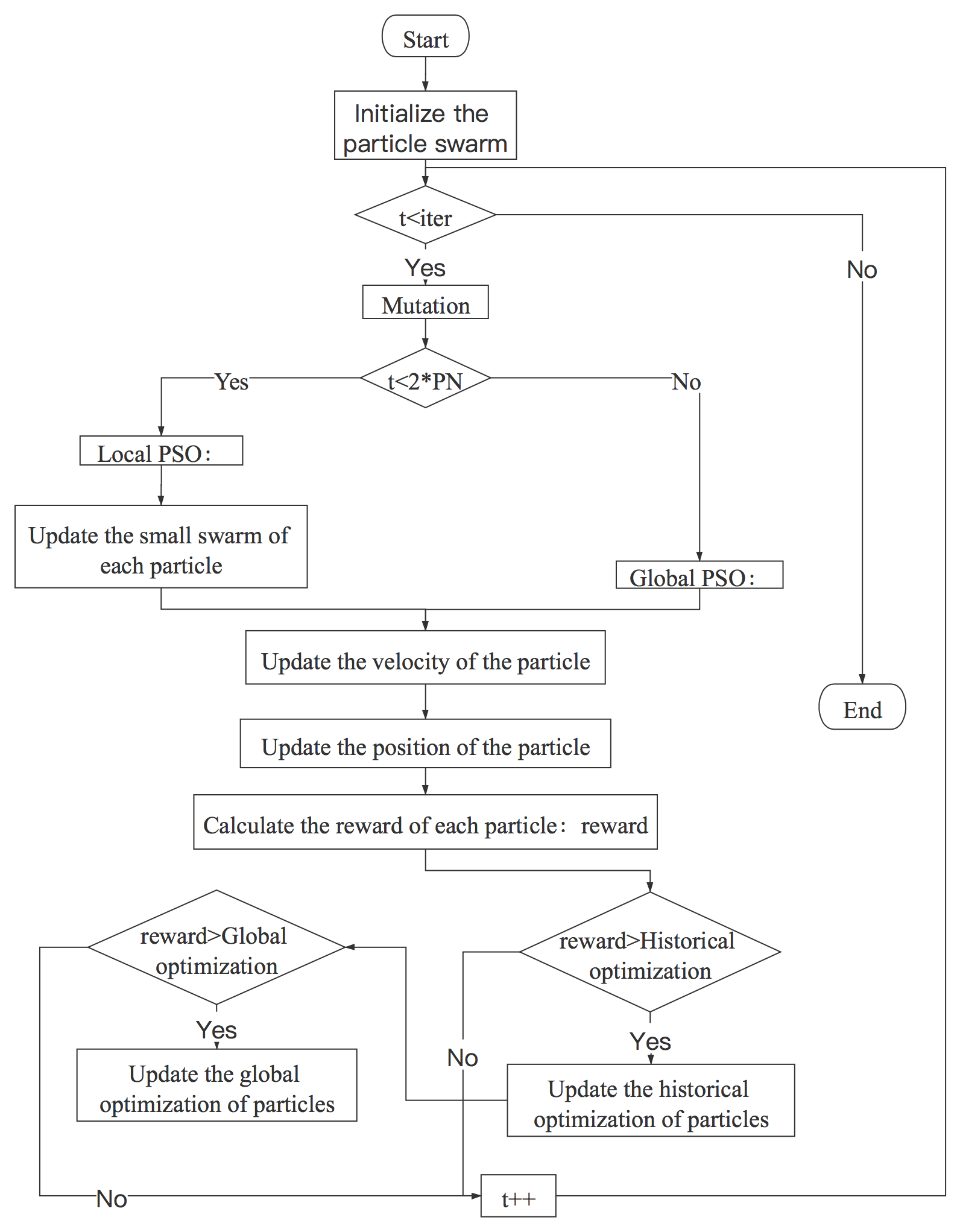}}
	\caption{Flow Chart of PSO}
	\label{fig5}
\end{figure}

The flow chart of PSO is shown in Fig.~\ref{fig5}. The first step is to initialize the particle swarm according to UAV number $|K|$ and target number $n$, which includes the initialization of the number of particles and iteration, the position of particle and the velocity of particle. In our design, $PN = 2(n+|K|-1)$ is the number of particles and $iter = 40(n+|K|-1)$ is the number of iterations. Both the position and the velocity of particle swarm are set to be $PN(n+|K|-1)$ dimensional arrays. Similar to GA described above, the first $n$ dimension of particle position represents the arrangement of targets, and the last $|K|-1$ dimension represents the way of dividing the targets. 

Secondly, in the mutation part, there is a probability that the particle position will change. Referring to \cite{b13}, the mutation probability of each iteration is set as 0.4, the particle number proportion of each mutation is set as 0.5, and the mutation position ratio of each mutation particle is set as 0.5. Thirdly, we use local PSO, in which all particles are divided into small swarms and the optimization is done separately in all small swarms, to jump out the local maximum in the early period. And then, in the velocity updating part, the new velocity of each particle is generated according to the current global optimal particle position and historical optimal particle position\cite{b14}. Then, in the position updating part, the new position of each particle is updated by the current position plus the new velocity. 

Then the reward of each particle is calculated. The reward is set as the total reward obtained by a particle.If the reward is greater than historical optimal solution, the historical optimal particle position will be updated to the current particle position and then if the reward is even greater than global optimal solution, the global optimal particle position will also be updated to the current particle position. The termination condition is when the number of iterations reaches the upper limit $iter$.

\section{Experiment and Result Analysis}
\subsection{Experiment settings}
Generally, intelligent algorithms cannot obtain global optimum solution and has a certain degree of randomness. To evaluate different algorithms fairly, a series of repeated experiments have been conducted. 

The experiments are divided into three groups, small scale, medium scale and large scale. Different groups have different settings, shown as Table \ref{tab1}. Except the number of UAVs and targets , other key parameters of the extended TOP are generated randomly, such as target positions, target rewards, time consumption at different targets and flight speeds. For one scale, 10 groups of parameters are generated randomly. Under each parameter setting, each algorithm solves 10 times. Intel Core i5-8250 CPU is used in the experiment.

\begin{table}[htbp]
	\caption{Experiment settings for different scales}
	\begin{center}
		\begin{tabular}{|c|c|c|c|}
			\hline
			\textbf{} & \textbf{\textit{Small scale}}& \textbf{\textit{Medium scale}}& \textbf{\textit{Large scale}} \\
			\hline
			UAV number&5&10&15 \\
			\hline
			Target number&30&60&90 \\
			\hline
		\end{tabular}
		\label{tab1}
	\end{center}
\end{table}

\subsection{Result Analysis}

\begin{figure}[htbp]
	\centerline{\includegraphics[width=9cm]{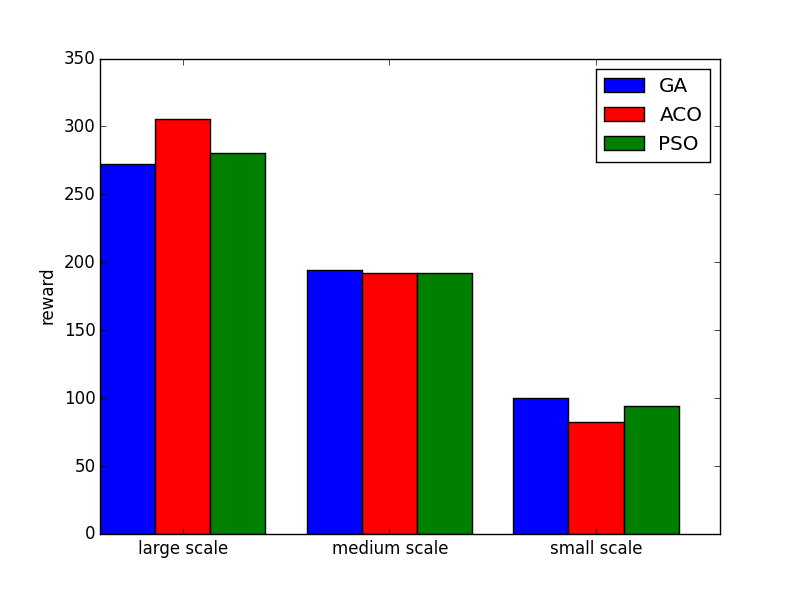}}
	\caption{Mean reward comparison among three algorithms}
	\label{fig6}
\end{figure}

\begin{figure}[htbp]
	\centerline{\includegraphics[width=9cm]{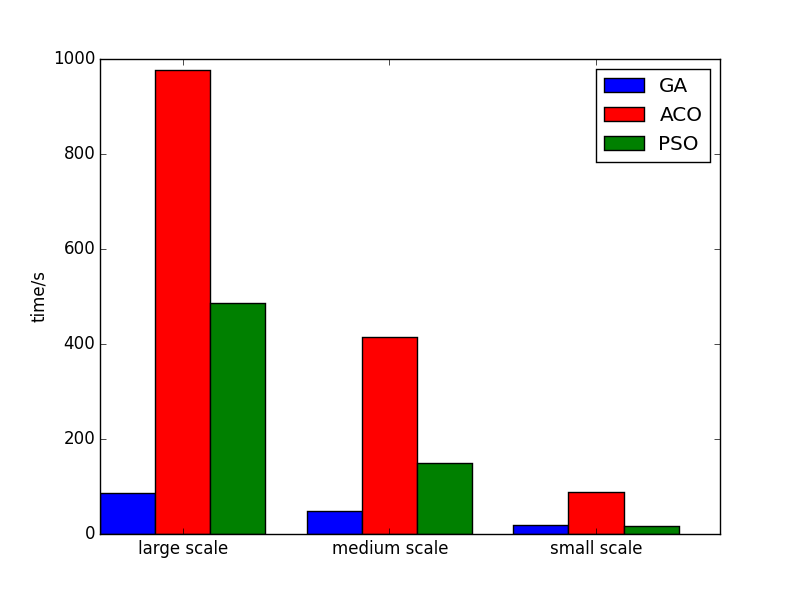}}
	\caption{Mean computational time usage comparison among three algorithms}
	\label{fig7}
\end{figure}

The evaluation index includes obtained reward and time complexity. The experiment results are shown as Fig.~\ref{fig6} and Fig.~\ref{fig7}. For mean reward, ACO performs best in the large scale group, but performs worst in the small scale. As a whole, three algorithms obtain similar rewards. However, for mean computational time usage, three algorithms have different performances: GA performs best, PSO follows, and ACO performs worst. Considering both obtained reward and time complexity, GA is recommended to solve the extended TOP among the three algorithms.


\section{Conclusion}
An extended Team Orienteering Problem is modeled for multi-UAV task assignment. Three intelligent algorithms, Genetic Algorithm, Ant Colony Optimization and Particle Swarm Optimization are implemented to solve the problem. A series of experiments are conducted and results are analyzed. The extended Team Orienteering Problem and experiment results constitute a benchmark for multi-UAV task assignment, which can be used to evaluate other algorithms.

\end{document}